\documentclass{article}

\usepackage{arxiv}
\usepackage{hyperref}       
\usepackage{url}            
\usepackage{booktabs}       
\usepackage{amsfonts}       
\usepackage{nicefrac}       
\usepackage{microtype}      
\usepackage{lipsum}
\usepackage{graphicx}
\usepackage{xcolor}
\usepackage{caption}
\usepackage{multirow}
\usepackage{amsmath}
\usepackage{makecell}
\usepackage{soul}
\usepackage[export]{adjustbox}

\definecolor{spcateg}{rgb}{0.77, 0.88, 0.71}
\definecolor{spevent}{rgb}{0.74, 0.84, 0.93}
\definecolor{spdriver}{rgb}{1, 0.9, 0.6}
\definecolor{spimpact}{rgb}{0.97, 0.79, 0.68}
\newcommand{\hlc}[2][yellow]{{%
    \colorlet{foo}{#1}%
    \sethlcolor{foo}\hl{#2}}%
}

\hypersetup{
pdftitle={Synthetic Text Generation using Hypergraph Representations},
pdfauthor={Natraj Raman}
}

\title{Synthetic Text Generation using Hypergraph Representations}

\author{
 Natraj Raman$^{1}$ and Sameena Shah$^{2}$ \\
  JPMorgan AI Research\\
  $^{1}$London, UK. \\
  $^{2}$New York, USA. \\
  \texttt{first.last@jpmorgan.com} \\
}

\begin{document}
\maketitle
\begin{abstract}
Generating synthetic variants of a document is often posed as text-to-text transformation.
We propose an alternate LLM based method that first decomposes a document into semantic frames and then generates text using this interim sparse format.
The frames are modeled using a hypergraph, which allows perturbing the frame contents in a principled manner. 
Specifically, new hyperedges are mined through topological analysis and complex polyadic relationships including hierarchy and temporal dynamics are accommodated.
We show that our solution generates documents that are diverse, coherent and vary in style, sentiment, format, composition and facts.
\end{abstract}

\section{Introduction}
Synthetic text plays a vital role in data augmentation, model robustness, privacy preservation and scenario analysis. It is usually formulated  as conditional text generation where a given source document is transformed using substitutions, paraphrasing, back translation, mixups etc. ~\cite{li2022data} to obtain a modified document.

We argue that conditioning on the unstructured text limits the ability to mix  text fragments coherently and produces transformations that are not confined to essential information, a critical necessity for long-form text. Furthermore, explaining the generated text becomes challenging, particularly detecting hallucinations ~\cite{filippova2020controlled}.

We propose here a \emph{decompose and expand} technique to generate synthetic text, where the semantic frames ~\cite{fillmore2006frame} of a source document are first extracted, and this compact interim form is used to generate the transformed text. These semantic frames capture only information relevant to target transformation and facilitate easier attribution of the generated text. While semantic parsing has been traditionally label intensive, we show that a Large Language Model (LLM) ~\cite{brown2020language} can be exploited in an in-context learning setting to obtain the frame elements with very few examples. The sparse semi-structured frames along with control instructions are then fed again into an LLM to generate text variants.  

A naive combination of the frame elements to introduce diversity may cover unrelated sources and produce incoherent documents. To address this, we follow a principled approach based on \emph{hypergraph networks} ~\cite{bretto2013hypergraph} to identify plausible combinations. In detail, each frame element is treated as a vertex and a frame as an hyperedge that can connect multiple vertices. New hyperedges are constructed by traversing the vertices based on their similarity in the feature space and the neighborhood topology. By augmenting with the frames corresponding to these new hyperedges, novel variations in the generated text is introduced.

Our hypergraph formulation provides a powerful mechanism that goes beyond classical dyadic relationships and models multi-way interactions among the semantic frames. This allows systematic representation of higher-order information and incorporating complex structures such as hierarchy and temporal overlap ~\cite{neuhauser2021consensus}. 

\begin{figure*}[tbp]
\centering
\includegraphics[width=\textwidth]{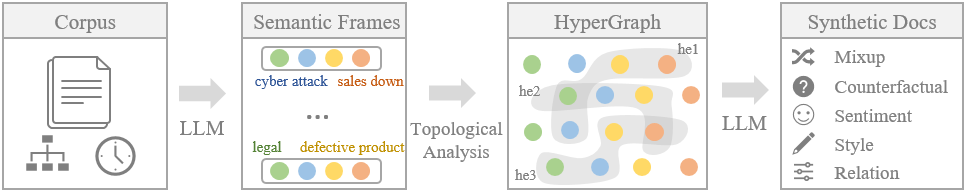}
\caption{Documents are converted into semantic frames (each dot is a frame element). New frames are constructed with hypergraphs and synthetic text with required variations is generated. An LLM is used for all text processing.  } 
\label{fig_overview}
\end{figure*}

We illustrate the efficacy of our technique with real-world documents from the financial domain to produce synthetic risk reports that vary in style, sentiment, format, composition and facts, with the added support for hierarchical and sequential relationships in a corpus. Fig. ~\ref{fig_overview} provides an overview.

Our core contributions are two-fold: (i) an LLM formulation for conditional text generation that uses intermediate semi-structured frames, and (ii) a hypergraph representation for interpolating higher-order interactions between semantic frames.

\section{Related Work}
Synthetic text is commonly created by altering tokens, sentences, fragments, or embeddings through rules, templates and language models. In data augmentation context they have been used for a variety of tasks, including sequence tagging, summarization, dialogs, and even multimodal data. See surveys ~\cite{feng2021survey, shorten2021text, chen2023empirical}. They have also been employed for adversarial scenario analysis ~\cite{madaan2021generate} and privacy preserving transformations ~\cite{sousa2023keep}.
Our work differs from typical approaches with its focus on compositional augmentation of long-form text, while utilizing interim structures and LLMs.

Recent efforts have exploited LLMs for semantic parsing ~\cite{drozdov2022compositional}, sequence label generation~\cite{liu2023gpteval}, structured output synthesis~\cite{josifoski2023exploiting}, creating questionnaire responses~\cite{hamalainen2023evaluating}, counterfactual generation ~\cite{fryer2022flexible} and differential privacy~\cite{kurakin2023harnessing}.  Despite sharing some LLM characteristics with these works, the use of frame concepts and hypergraph mining differentiates our effort.
\section{Model}
\noindent \textbf{Semantic Parsing}: Given a corpus of $N$ documents $\{d^i\}_{i=1}^N$, we wish to produce synthetic variants of these documents. We first subject each $d^i$ to semantic parsing using an LLM with few shot prompting and obtain $M_i$ different semantic frames $\{\mathbf{h}^i_m\}_{m=1}^{M_i}$, where each frame has $K$ different frame elements, i.e. $\mathbf{h}^i_m=(h^i_{m1}...h^i_{mK})$. A frame element is usually a text phrase and the frames compactly encode the original text.

\vspace{0.2cm}
\noindent \textbf{Hypergraphs}: The frames are represented using a hypergraph to facilitate the mining of new frames. Let $\mathcal{G}=(\mathcal{V},\mathcal{E})$ be an undirected heterogeneous hypergraph, where each frame element $h^i_{mk} \in \mathcal{V}$ is a node and a frame $\mathbf{h}^i_m \in \mathcal{E}$ is an hyperedge. $\mathcal{G}$ is K-uniform and has same degree of heterogenity across all of $\mathcal{E}$. Each node has feature vectors $x^i_{mk} \in \mathbb{R}^p$. Here the embeddings corresponding to a frame element text is used as its feature.   

\vspace{0.2cm}
\noindent \textbf{Hyperedge Mining}: We define the relationship strength between any two hyperedges using a Gaussian Kernel as 
\begin{align}
    A(i_m, j_{m'}) = \sum_{k=1}^K exp\left(-\gamma\delta(x^i_{mk},x^j_{m'k})\right),
\end{align}
where $\delta:(x,y)\to\mathbb{R}$ is a distance function, $\gamma$ is a bandwidth parameter and  $A \in \mathbb{R}^{|\mathcal{E}|\times|\mathcal{E}|}$ captures the inter-hyperedge strength. To account for higher-order relations, we perform a link analysis based on PageRank ~\cite{gleich2015pagerank} algorithm with a damping factor $\alpha \in [0,1]$, a normalized $\bar{A}$ and obtain an intimacy matrix $S$ as
\begin{align}
S = \alpha\bar{A} + (1-\alpha)1/|\mathcal{E}|.
\end{align}
The topk hyperedges that are most intimate to an hyperedge $i_m$ is identified from $argsort(S(i_m,:))$. In practice, only inter-document frames and hyperedges within an $\epsilon$-ball neighborhood is considered. Two intimate hyperedges ($\mathbf{h}^i_m$,   $\mathbf{h}^j_{m'}$) are mixed as 
\begin{align}
\mathbf{h}^{i*}_m & = \mathbf{h}^i_m\odot b^* \oplus \mathbf{h}^j_{m'}\odot (1-b^*) \nonumber \\
\mathbf{h}^{j*}_{m'} & = \mathbf{h}^i_m\odot (1-b^*) \oplus \mathbf{h}^j_{m'}\odot b^*,
\end{align}
where $b^*\in\{0,1\}^K$ is a binary vector sampled based on $\delta$, and $\odot, \oplus$ are frame combination operators. These new hyperedges $\mathbf{h}^*$ are added to $\mathcal{G}$, thereby increasing the number of hyperedges while the number of vertices remain fixed.

\vspace{0.2cm}
\noindent \textbf{Hierarchical Relations}: Let $W \in \mathbb{R}^{N\times N}$ be inter-document similarity weights, known apriori. For instance, documents $i,j$ that share the same hierarchical level might have higher $W^{ij}$.  This relationship can be directly incorporated by scaling (1) with $W^{ij}$, thereby biasing the frame mixups in (3) towards documents that are related in hierarchy.

\vspace{0.2cm}
\noindent \textbf{Temporal Dynamics}: Documents often evolve over time and by distinguishing between static and dynamic frames, we can generate better text variants. To accommodate a time instance $t \in {1...T}$, let $\mathbf{h}^i_{mt}$ be a frame now. We extend (1) to capture time dependent strengths $A(i_{mt}, j_{m't'})$ and define the history $\mathbb{T}$ of a frame  as 
\begin{align}
\mathbb{T}(\mathbf{h}^i_{mt}) = \{t'| A(i_{mt}, j_{m't'}) > \epsilon,  \forall t'\scriptstyle =1...T \displaystyle \},
\end{align}
where $\epsilon$ is a similarity threshold. This history can be used during text generation to compare and contrast documents over a timeline. 

\vspace{0.2cm}
\noindent \textbf{Text Generation}: The existing frames of a document along with the newly mined frames are fed into an LLM to generate the surface realizations. Prompt instructions to the LLM can differ based on a desired control attribute such as a preferred style or format, and thus we can produce multiple variants for the same document. 
\section{Experiments}
\label{sec_results}

\subsection{Setup}

\noindent \textbf{Dataset}: We conduct experiments on the financial reports dataset ~\cite{financial-reports-sec}, which comprises of  annual 10-K reports from publicly traded US companies. Each document in this dataset can run into multiple pages with several sections, and we focus on generating synthetic versions of the risk section (i.e. Item 1A). The first 4056 tokens present in this risk section for over 640 different companies is utilized as the original text in our experiments. Table ~\ref{tab_origtext} in the Appendix contains few sample excerpts from the dataset.

\noindent \textbf{Settings}: We use GPT4 ~\cite{gpt42023} as the LLM with a temperature value of $0$. We set $K=4$, $\gamma=1$, $\alpha=0.85$, $\epsilon=0.8$ and use cosine distance for $\delta$. 

\vspace{0.2cm}
\noindent \textbf{Semantic Frames}: We define the semantic frame elements as risk category (e.g. legal, credit), event (e.g. customer bankruptcy), driver (e.g. market instability) and impact (e.g. reduced revenue). The elements can contain any free-text. Table ~\ref{tab_semparse} contains examples of the original text and their parsed semantic frames. Table ~\ref{tab_semparseprompt} contains the LLM prompt.

\vspace{0.2cm}
\noindent \textbf{Evaluation Baselines}: We compare the text generated with our frame hypergraph method against popular baselines BERT substitution ~\cite{kenton2019bert},  BART paraphrasing ~\cite{lewis2020bart}, Marian back translate ~\cite{mariannmt} and T5 ~\cite{raffel2020exploring} and GPT4 summarizer.  For contextual word embedding based substitutions, we use ~\cite{ma2019nlpaug}. A publicly available model\footnote{https://huggingface.co/eugenesiow/bart-paraphrase} that was trained on Quora, PAWS and MSR paraphrase corpus is used for BART paraphrasing. The back translation model employed translation from English to French followed by French to English and is accessible\footnote{https://huggingface.co/Helsinki-NLP/opus-mt-en-roa}. The T5 summarizer model was trained on news articles and is publicly available\footnote{https://huggingface.co/mrm8488/t5-base-finetuned-summarize-news}. A consistent text length is used for all these language models and the results are reported for a single run using a fixed seed. These baselines cover a wide range of approaches with differing capability, complexity, stability and quality levels. 

\vspace{0.2cm}
\noindent \textbf{Evaluation Metrics}: The grammatical acceptability of the generated text is assessed through the fluency metric ~\cite{style20}. We use a RoBERTa-large model trained on the CoLA corpus\footnote{https://huggingface.co/cointegrated/roberta-large-cola-krishna2020} for this purpose. We also assess the similarity and diversity between the generated and original text passages. BLEU, GloVe and SentenceTransformer ~\cite{reimers-2019-sentence-bert} embeddings are used for lexical, semantic word and semantic sentence similarity metrics respectively. The GloVe and SentenceTransformer embeddings comparison is based on cosine similarity. For divergence, we use distinct n-grams and mean embedding similarity analogous to ~\cite{tevet2021evaluating}.  Furthermore, we rate the degree of semantic similarity between between the original and generated text using the coherence metric. We particularly focus on the new sentences that were introduced as a result of mixups and compare them with then an equivalent number of sentences from the original text using the SentenceTransformer embeddings\footnote{https://huggingface.co/sentence-transformers/all-MiniLM-L6-v2}.

\begin{table*}[tbp]
\caption{Sample excerpts from the financial reports dataset ~\cite{financial-reports-sec}. These original text are used for semantic parsing and comparing with synthetic variations. }
\centering
\small
\begin{tabular}{|p{\linewidth}|}
\hline
The following is a description of some of the principal risks inherent in our business. The risks and uncertainties described below are not the only ones facing us. We may be affected by continuing problems in the aviation industry. As a provider of products and services to the aviation industry, we are greatly affected by the overall economic condition of that industry. The aviation industry is historically cyclical. <snipped>  \\
 \\
\hline
 Our business is subject to numerous risks, including but not limited to those set forth below. Patent litigation is very expensive and we may not have sufficient cash available to pursue any patent litigation to its conclusion because currently we do not generate revenues. We are dependent upon the success of our patent infringement lawsuits. If we lose our key personnel our operations could be harmed. <snipped>  \\ 
\\

\hline
\end{tabular}
\label{tab_origtext}
\end{table*}

\begin{table*}[tbp]
\caption{Example results of the semantic parser with input text and output frames. The frame elements are risk \hlc[spcateg]{category}, \hlc[spevent]{event}, \hlc[spdriver]{driver} and \hlc[spimpact]{impact}.    }
\centering
\small
\begin{tabular}{|p{\linewidth}|}
\hline
 As a result of our continuing review of our business, we may have to undertake further restructuring plans that would require additional charges, including incurring facility exit and restructuring charges. \\
 \\
 \hlc[spcateg]{operational} \hlc[spevent]{restructuring plans} \hlc[spdriver]{ongoing business review} \hlc[spimpact]{additional charges} \\
 \hline
 Our supply of products is derived from only one vendor and there can be no assurance that the vendor would continue to, or have the ability to, continue supply of product. Should the vendor discontinue supplying product, we would lose 100\% of its supply of product.  \\ 
 \\
 \hlc[spcateg]{supplychain} \hlc[spevent]{supply of product} \hlc[spdriver]{single vendor} \hlc[spimpact]{loss of 100\% of product supply} \\
\hline
 We have no operations and minimal assets, which raises substantial doubt about our ability to return any additional value to our stockholders. \\ 
 \\
 \hlc[spcateg]{credit} \hlc[spevent]{no operations and minimal assets} \hlc[spdriver]{n/a} \hlc[spimpact]{inability to return value to stockholders} \\

\hline
\end{tabular}
\label{tab_semparse}
\end{table*}

\begin{figure*}[tbp]
\centering
\includegraphics[width=\textwidth]{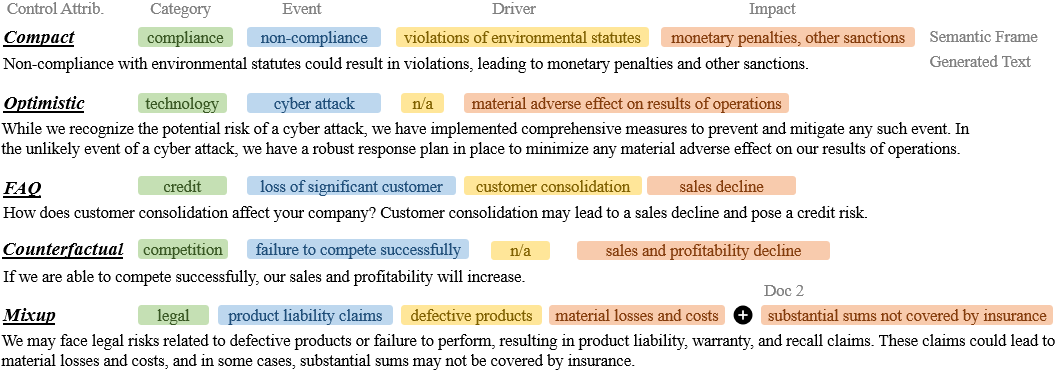}
\caption{Qualitative examples of text generated from semantic frames that were parsed from a risk report. They correspond to 5 different control attributes and illustrate variations in style, format, composition and facts.  } 
\label{fig_qual}
\end{figure*}

\begin{table*}[!tbp]
\caption{Comparison results for conditional text generation methods using fluency, similarity and diversity metrics. }
\centering
\small
\begin{tabular}{|l|r|r|r|r|r|r|r|}
\hline
 \multicolumn{1}{|l|}{}&\multicolumn{1}{|r|}{}&\multicolumn{3}{c|}{Similarity}&\multicolumn{3}{c|}{Diversity} \\
 \cline{3-5} \cline{6-8}
  Method  & Fluency & Lexical & SemWord & SemSent  & Lexical & SemWord & SemSent \\
  \hline
BERT Substitution          & 54.1 & 46.7 & 90.0 & 78.8 & 44.0 & 16.2 & 31.8\\
BART Paraphrase            & 93.4 & 93.1 & 99.5 & 97.6 &  5.2 & 16.2 &  6.1\\
Marian Backtranslate       & 85.1 & 65.5 & 95.7 & 88.5 & 27.8 & 17.2 & 20.3\\
T5 Summarize               & 93.8 & 85.3 & 99.2 & 82.3 & 10.3 & 19.1 & 32.7\\
GPT4 Summarize             & 98.1 & 49.4 & 96.3 & 78.3 & 36.8 & 15.9 & 31.7\\
GPT4 + Frames              & 97.9 & 41.3 & 94.3 & 71.6 & 41.9 & 18.7 & 41.8\\
GPT4 + FrameMixups         & 98.0 & 38.0 & 92.8 & 68.7 & 43.4 & 20.0 & 46.2\\

\hline
\end{tabular}
\label{tab_rescompare}
\end{table*}

\begin{figure}[!tbp]
\centering
\includegraphics[height=3.5cm]{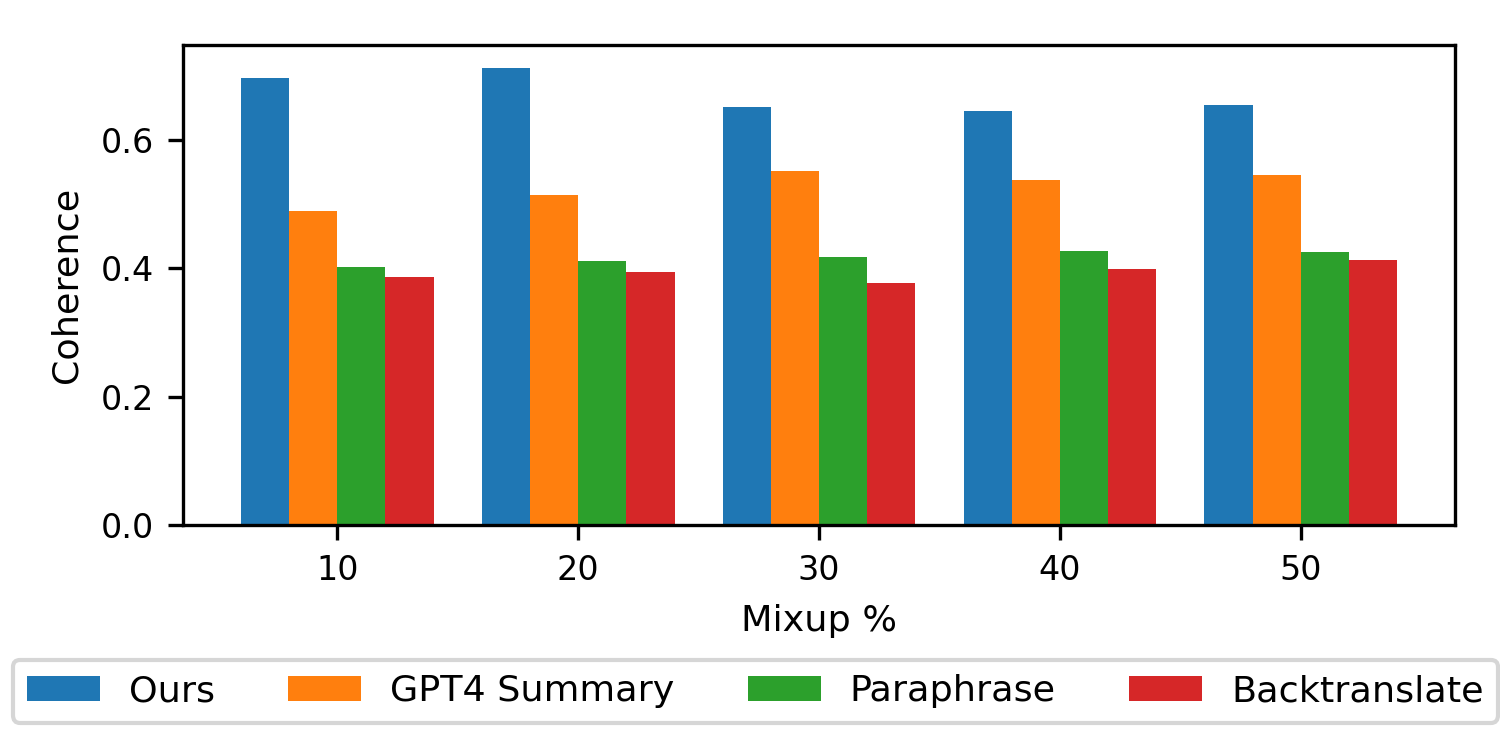}
\caption{Comparative analysis of coherence as new documents are mixed to the original in different ratios.  } 
\label{fig_mixins}
\end{figure}

\subsection{Results}
\vspace{0.2cm}
\noindent \textbf{Text Generation}: Table ~\ref{fig_qual} contains qualitative examples synthetic text that were generated directly from the semi-structured semantic frames. They correspond to five different control attributes that are used to tailor the conditional generation outputs and Table ~\ref{tab_genprompt} has few sample prompts. The \emph{Compact} format simply develops the frames into fluent text. The \emph{Optimistic} sentiment adds context to minimize the risk impact. The \emph{FAQ} style produces output in a question answer interface. The \emph{Counterfactual} setting presents an adversarial scenario. Finally, the \emph{Mixups} demonstrate the combination of frames from other documents. The key point to observe here is the development of sparse incomplete information into fluent content and the variations in style, tone, format and facts as dictated by the control attributes. Note that the output shown is only an excerpt and in practice the generated text is much longer and correspond to several frames. Thus we can generate multiple alternative surface realizations for a single set of frames or from a sampled subset of frames. This frame based approach treats information relevant to text generation as a cohesive unit and offers considerable advantage over classical conditional text generation solutions where sentences or fragments maybe randomly sampled, ignoring vital linguistic connections.

\vspace{0.2cm}
\noindent \textbf{Quantitative Comparison}:
Table ~\ref{tab_rescompare} shows the comparison results across various baselines.  The chosen baselines reflect common existing approaches for synthetic text generation and also includes a state-of-the-art LLM, GPT4. It is unsurprising that GPT4 based models score extremely well in fluency and that their term based similarity are low while their word embedding similarity is high, denoting their rich vocabulary. The semantic similarity for GPT4 models are comparable, but drops when using the frames due to their sparse nature and when new information is added during document mix up. In contrast, the diversity for frame models increase, indicating considerable variations from original text.  Although mixing different frames can increase diversity, the generated text must maintain a logical flow and a consistent focus on the central theme.  Fig. ~\ref{fig_mixins} shows a comparison of coherence among various methods as new information is added in varying proportions. It is observed that even as we introduce new information from other documents (i.e. new frames in our approach, while new sentences in the baselines), the proposed solution produces text that is more coherent than the alternatives. The above results illustrate that our semantic frame hypergraph method produces text that is both diverse and coherent. 

\noindent \textbf{Discussion}:
The primary focus of the quantitative comparison is to verify whether our solution can improve the synthetic text generation process even for an exceptionally powerful model such as GPT4. Despite the sparse nature of frames and mixing frames across documents, the semantic similarity between the original and generated text is comparable to other methods. Importantly, our evidence shows that the frame based approach produces new text that is more diverse and yet related to the original text. Furthermore, the coherence plot confirms that the degree of semantic relatedness in generated content is preserved when documents are mixed. The key insight from these results is that rather than a monolithic approach to conditional text generation, the decompose and expand technique proposed here can exploit the LLMs in a better manner and the hypergraph formulation to mine new frames for information mixup results in coherence preserving text.

\begin{figure}[!tbp]
\centering
\includegraphics[height=4cm]{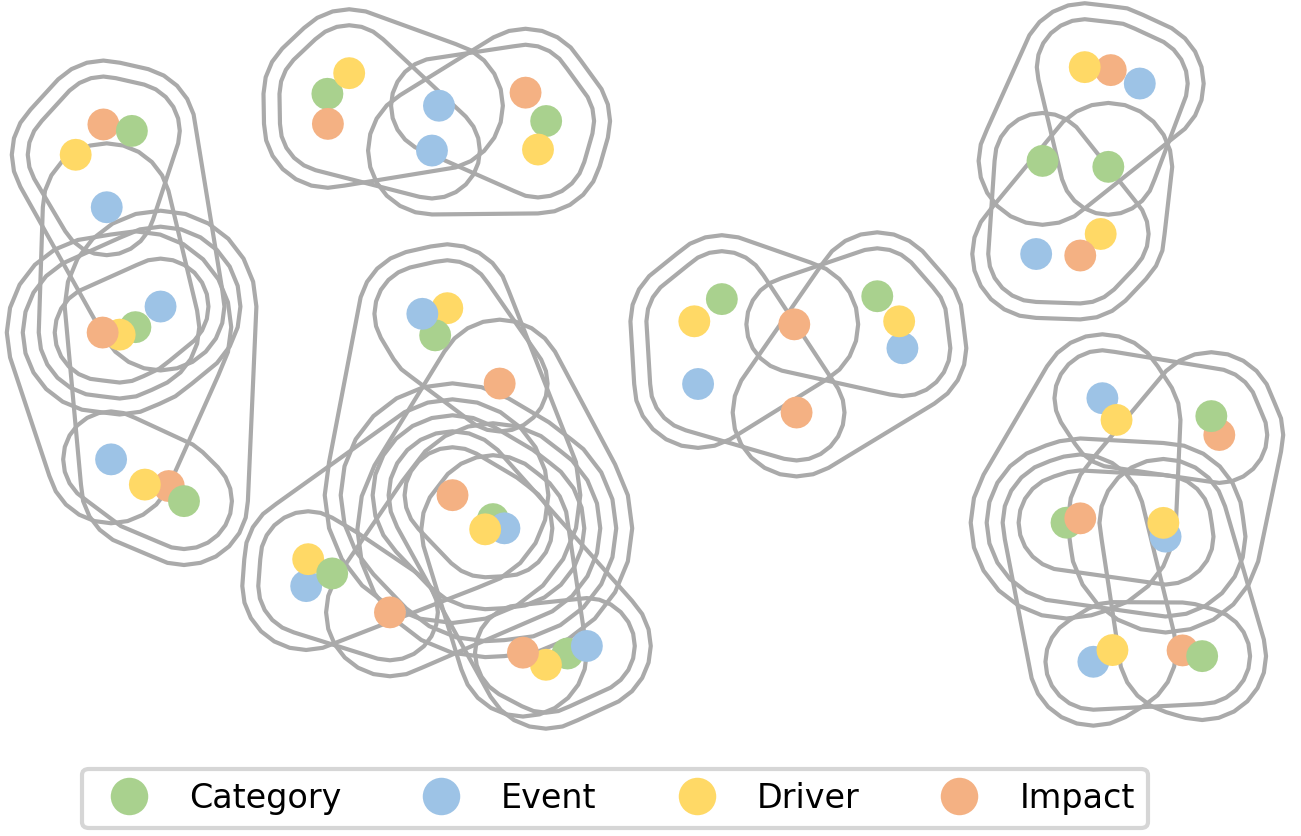}
\caption{Hyperedge visualization. Frame elements (dots) span multiple frames to create rich mixups.  } 
\label{fig_heviz}
\end{figure}

\begin{figure}[!tbp]
\centering
\includegraphics[height=4.6cm]{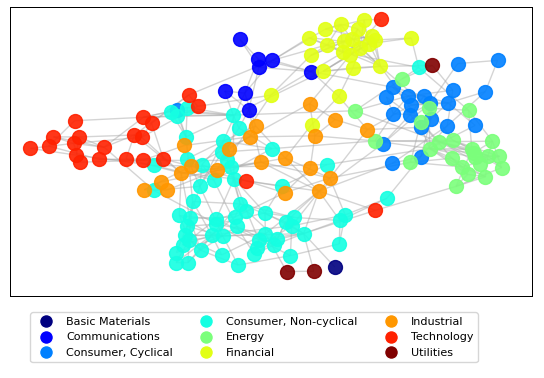}
\caption{Network homophily when including hierarchical relations - similar sectors appear connected.  } 
\label{fig_hiermixins}
\end{figure}

\begin{figure}[!htbp]
\centering
\includegraphics[height=4.6cm]{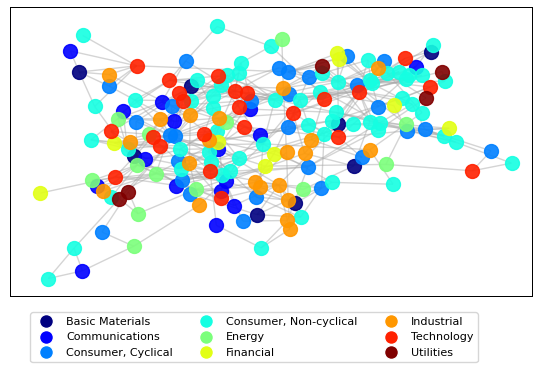}
\caption{Frames across different sectors appear connected when hierarchical relations are not included.  } 
\label{fig_nohiermixins}
\end{figure}

\begin{figure}[!htbp]
\centering
\includegraphics[height=4.6cm]{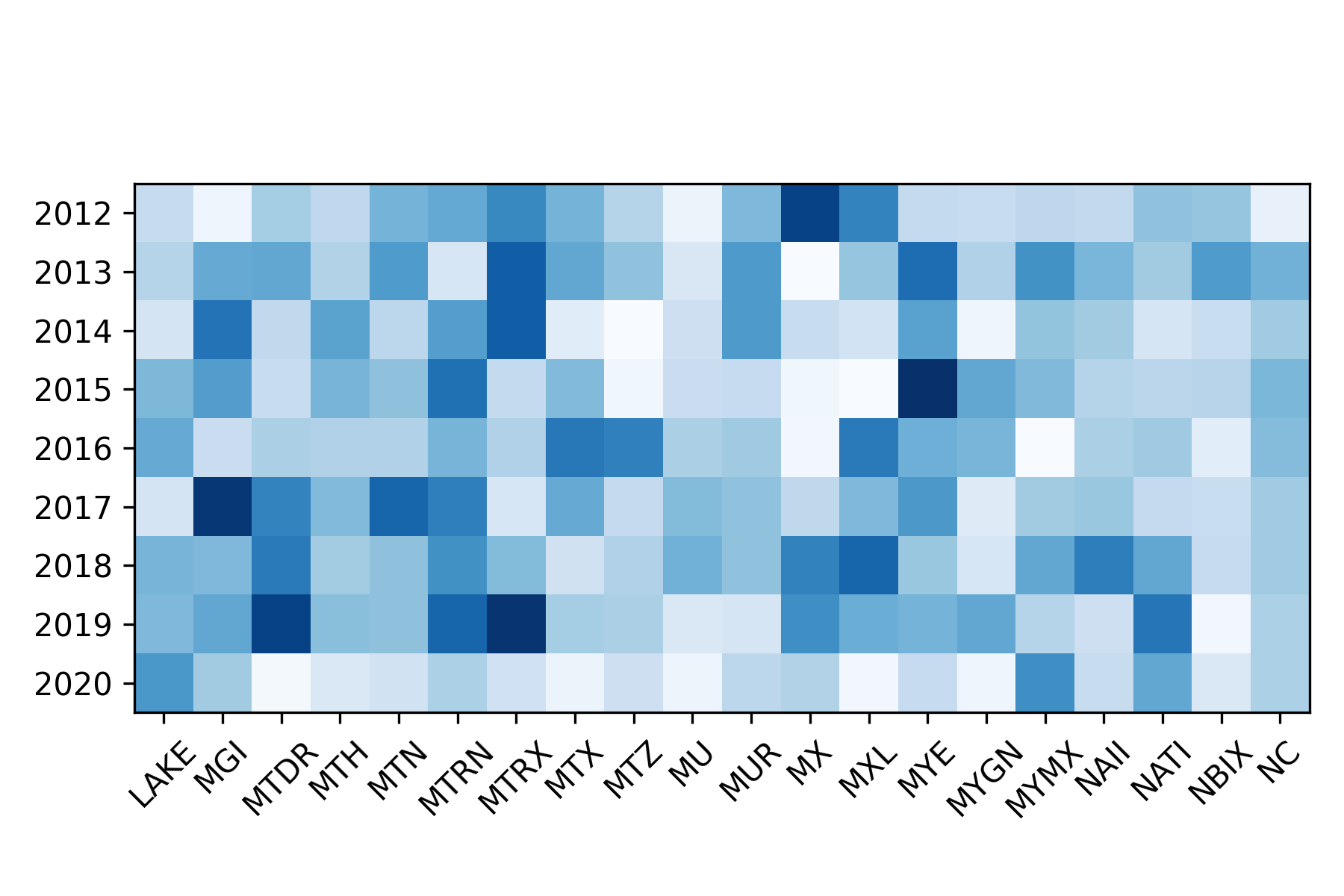}
\caption{Heatmap visualization of historical similarities between semantic frames. It highlights the volume of new risks observed in a company's risk report, over a ten year period. } 
\label{fig_temporal}
\end{figure}

\begin{table}[!tbp]
\caption{Synthetic reports with temporal summaries showing historical and emerging risks. }
\centering
\small
\begin{tabular}{p{\linewidth}}
\Xhline{3\arrayrulewidth}
The ongoing COVID-19 pandemic ... . This is a \emph{new risk that has emerged recently} and requires our attention. \\
\\
Operational risks have been a concern \emph{in the past}, as evidenced by our reports from 2017, 2018, and 2019.  \\
\\
We \emph{continue to face} these risks in the current year.\\

\Xhline{3\arrayrulewidth}
\end{tabular}
\label{tab_temporal}
\end{table}

\subsection{Structure Analysis}
\vspace{0.2cm}
\noindent \textbf{Frame Combinations}:
When constructing the frame mixtures, it is desirable to mix over several frames instead of focusing on only pairwise combinations. Fig. ~\ref{fig_heviz} visualizes a slice of the hypergraph constructed after frame mixups, with the frame elements (i.e. company, event, driver and impact) being represented as colored circles and their combinations shown as gray lines. It highlights that the frame mixtures are inter-linked over several frames, resulting in multifaceted combinations. 

\vspace{0.2cm}
\noindent \textbf{Frame Hierarchies}:
Documents in a corpus may be hierarchically related, indicating that documents with a common parent share similar themes and concepts. When mixing frames across different documents, it is important to respect this hierarchy in order to ensure that the synthetically generated content discusses consistent themes. For instance, companies within the same industry sector hierarchy should be prioritized over those from a different industry sector when merging semantic frames, as their risk impacts maybe more comparable. However this prioritization is a soft preference, with the similarity between frames also a contributing factor to frame mixups. The network graph in Fig. ~\ref{fig_nohiermixins} plots the connections between frames (i.e. hyperedges) with each frame being colored by its industrial sector (i.e. hierarchy). The frames were mixed up ignoring hierarchical relations, and we see the heterogeneity in colors between any two connected nodes resulting in diffuse connections. In contrast, Fig ~\ref{fig_hiermixins} illustrates the connections when hierarchical relationship is included and we observe the desired trait of network homophily, with frames from similar sectors tending to form homogeneous connections. This illustrates the ability of our mixup technique to accommodate both frame similarities and document hierarchies.

\vspace{0.2cm}
\noindent \textbf{Frame Temporal Dynamics}:
It is useful to observe the extent of changes that documents undergo over time. For instance, companies publish risk reports over several years and it is important to identify what has changed across years to get a condensed summary. Table ~\ref{tab_temporal} contains few temporal summaries that were created from sequences of frames, where each frame is associated with the timestamp of its parent document. We can see that the summary captures both historical risks and new emerging risks and allows generating documents that track the evolution of recurring concepts and new themes. In addition to the summary, it is also useful to highlight the time period during which there are significant changes in content.  A naive comparison of the unstructured text is a flawed approach and can be misleading. By explicitly modeling the temporal information and using a semi-structured representation, temporal frame semantics provide an accurate mechanism to capture similarities and differences.  Fig. ~\ref{fig_temporal} displays a heatmap illustrating frame similarities calculated for multiple companies' risk reports over a ten year period and helps in visualizing the magnitude of new risks emerging each year. Thus the frame temporal dynamics helps in creating compact overviews, comparing previous years and identifying trends.  

\begin{table*}[tbp]
\caption{ Comparative evaluation of topological methods for mining new semantic frames.}
\centering
\small
\begin{tabular}{|l|r|r|r|}
\hline
 {Method} & {Document Diversity} & {Topic Diversity} & {Content Diversity} \\ 
  \hline
Jaccard Coefficient                 & 36.1 & 34.1 & 54.7 \\
Preferential Attachment             & 27.4 & 26.4 & 54.5 \\
Adamic Adar                         & 38.3 & 36.7 & 65.7 \\
Resource Allocation Index           & 39.1 & 36.9 & 65.3 \\
Common Neighbor Centrality      & 36.2 & 35.3 & 61.4 \\
Hypergraph (Ours)                   & 49.2 & 39.3 & 80.0 \\

\hline
\end{tabular}
\label{tab_hypcomp}
\end{table*}

\begin{figure}[tbp]
\centering
\includegraphics[height=5cm]{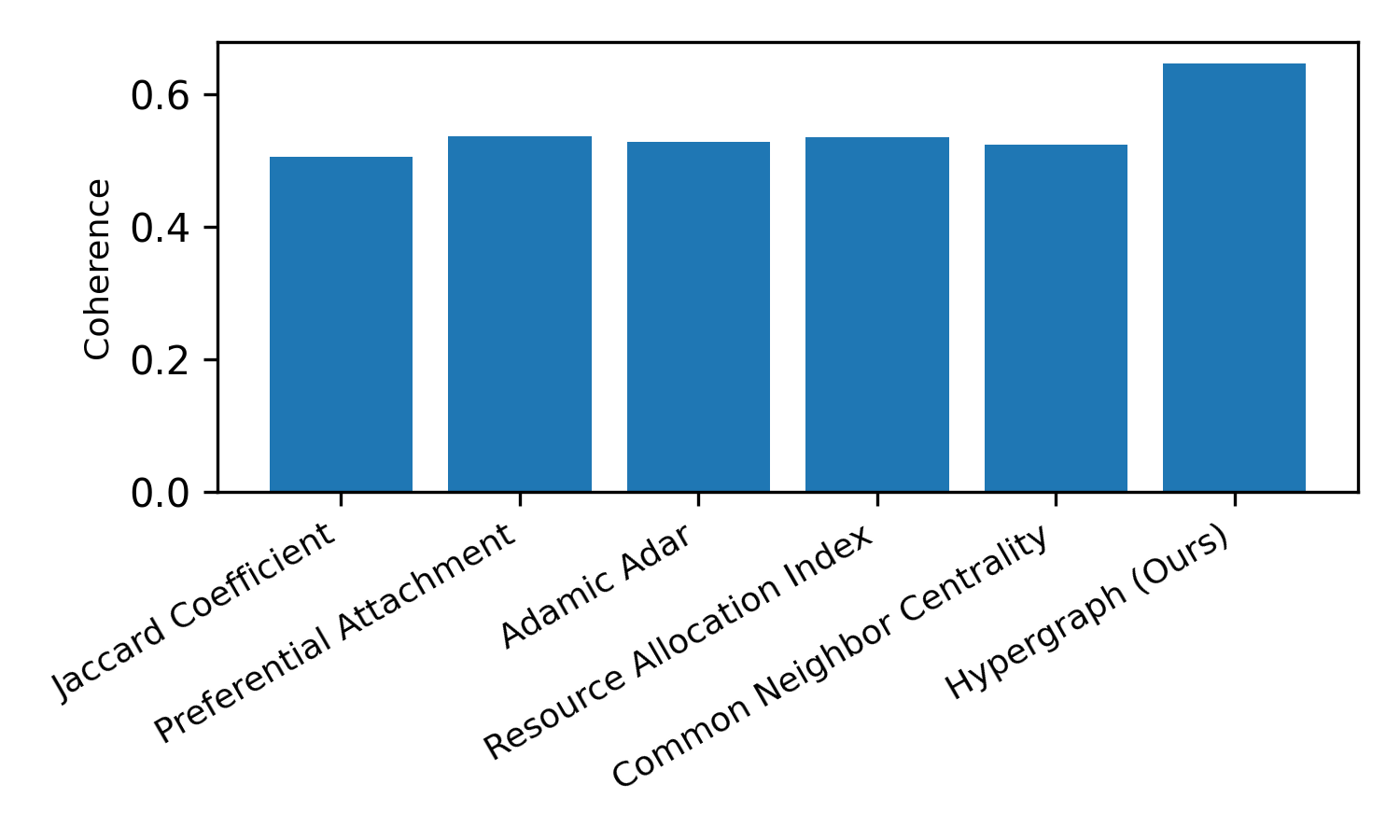}
\caption{Comparative analysis of coherence in generated content for different link prediction methods. All the methods used GPT4 for text generation.  } 
\label{fig_hypcohviz}
\end{figure}

\subsection{Additional Analysis}
\noindent \textbf{Why Hypergraphs?}:
We analyze the utility of hypergraph representations for combining the semantic frames as part of ablation study. In particular, we investigate whether simpler alternatives can introduce new semantic frames to an existing document equally well. We cast the problem of identifying the new frames to mix as a link prediction problem. Each frame is treated as a node in a traditional graph and if a new link is predicted between any two frames, then we use these frames to mix the documents. We employ several link prediction algorithms ~\cite{liben2003link} that analyze the node neighborhoods to produce a ranked list of probable edges. Specifically, Jaccard Coefficient, Preferential Attachment,  Adamic Adar, Resource Allocation Index and Common Neighbor Centrality~\cite{ahmad2020missing} are used.  We measure the extent of new documents  mixed into the original (document diversity) and the diversity in topics (i.e. risk category element) and content (i.e. risk event, driver and impact elements).   Table ~\ref{tab_hypcomp} compares these methods against our solution and we can observe that the hypergraph based topological analysis outperforms other solutions. Furthermore, we compare the coherence scores for these methods in Fig ~\ref{fig_hypcohviz}. All of these approaches employ GPT4 for text generation, with only the frame mixup methods exhibiting variation. Our solution produces more coherent text compared to both other language models (Fig ~\ref{fig_mixins}) and conventional link prediction algorithms.

\begin{figure*}[!tbp]
\centering
\includegraphics[width=\textwidth]{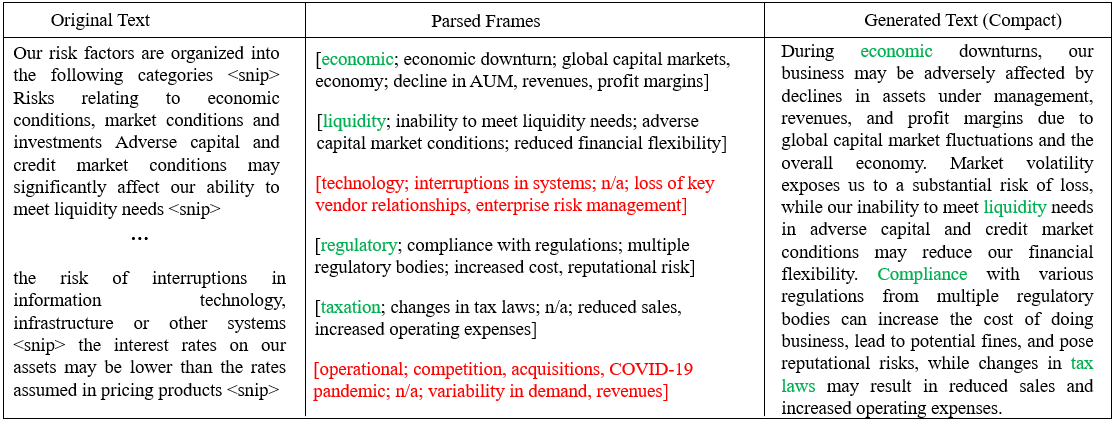}
\caption{Enhanced interpretability of the frame mechanism. The original text that is conditioned on, the intermediate frames parsed from the original and the text generated from the frames are shown. Frames are highlighted with green or red based on whether the information it contains was represented in the generated text.} 
\label{fig_interp}
\end{figure*}

\vspace{0.2cm}
\noindent \textbf{Interpretability}:
A notable benefit of employing semantic frames as an intermediary form lies in the enhanced interpretability it offers. During synthetic text generation, it is often required to discern the similarities and differences between the original and generated text to detect omitted information or unintended content. A direct comparison is challenging due to the unstructured nature of the textual content. It is much easier to compare the semi-structured frames with the generated sentences and identify the frames that are not present in the generated text or new text that do not correspond to any frame. We demonstrate this advantage in Fig ~\ref{fig_interp}, where we investigate a compact text generation setting in which the LLM is instructed to generate exactly three sentences. We highlight the information that was both present (green) and absent (red) in the parsed frames based on text similarity measures between the frames and the generated text. Although this exercise could be performed between the original and generated text, the lengthy format and redundant information in the original text make comprehending the modifications more complex. Thus the use of intermediate frames provides an elegant mechanism to explain the generated content. 

\section {Conclusion}
We presented a two stage approach for synthetic text generation: text to frames and frames to text. Our use of LLMs makes this modular few shot approach feasible. Importantly, our hypergraph formulation models complex frame interactions and thus helps generating documents that are both diverse and coherent. Our future work will focus on conversational text and generative hypergraphs.

\section*{Limitations}
The semantic frames vary with each dataset, and their definition requires domain knowledge. While a generic family of frame structure could be specified, the resulting ontology will be elaborate and the LLMs may struggle with in-context learning and therefore require sizable number of examples to finetune the semantic parser and the generator. Furthermore, our method assumes that the LLM parsing and generation is atomic. Complex memory management routines may be necessary when handling very large text that cannot be fit into an LLM context. 

The hypergraph formulation for mining new semantic frames introduces additional complexity. However, its generality, expressiveness  ~\cite{antelmi2023survey} and ability to handle many-to-many relationships, group interactions and temporal dynamics makes it a better choice than simpler adhoc solutions. 

New hyperedges are constructed purely based on the similarity between the frame elements. In practice, the mixups may be governed by complex constraints for joining the frames and avoiding infeasible combinations will require extensions to the mining procedure proposed here.      

\section*{Ethics Statement}
Our method relies on LLMs to produce synthetic text. Therefore, we are susceptible to any ethical concerns associated with LLMs~\cite{zhuo2023exploring}.  In addition, the text generated through frame mixups may amplify biases already present in the original text, ignore important nuances, misrepresent facts or introduce unexpected concepts. A subset of these effects may in fact be desirable, e.g. for validating the robustness of a trained model or understanding its limitations. Some of the impacts can also be dampened by selecting appropriate hyper parameter values, e.g. the mixup ratio. In general, awareness of the potential risks and the available choices can mitigate these concerns.

\section*{Acknowledgments}{This paper was prepared for information purposes by the Artificial Intelligence Research group of JPMorgan Chase \& Co and its affiliates (“JP Morgan”), and is not a product of the Research Department of JP Morgan.  J.P. Morgan makes no representation and warranty whatsoever and disclaims all liability for the completeness, accuracy or reliability of the information contained herein. This document is not intended as investment research or investment advice, or a recommendation, offer or solicitation for the purchase or sale of any security, financial instrument, financial product or service, or to be used in any way for evaluating the merits of participating in any transaction, and shall not constitute a solicitation under any jurisdiction or to any person, if such solicitation under such jurisdiction or to such person would be unlawful. © 2023 JP Morgan Chase \& Co. All rights reserved.}

\bibliographystyle{unsrt}  
\bibliography{references}  

\section*{Appendix}
\appendix

\section {Prompts}

\begin{table}[!htbp]
\caption{GPT4 prompt for parsing a text into semantic frames in an in-context learning setting.    }
\centering
\small
\begin{tabular}{|p{\linewidth}|}
\hline
For a given text passage, list the risk category (can be one of credit, market, liquidity, operational, compliance, regulatory, legal, capital, conduct, strategic, technology, reputation, supplychain, environment), the risk event (i.e. threat), the risk driver (i.e. underlying causes) and the risk impact (i.e consequence) in the form of a tuple [<category>; <event>; <driver>; <impact>]. There can be several such tuples in a passage. Some sentences maybe irrelevant and a tuple item may not be applicable (e.g. impact is not available).\\ 
\\
Example \#1\\
From time to time certain of our customers have filed for bankruptcy protection or ceased operation. The impact of instability in the global financial markets may lead to reduced domestic or international capacity. Future environmental regulatory developments in the United States and abroad concerning environmental issues, such as climate change, could adversely affect our operations and increase operating costs.\\
\\
Answer \#1\\
{[operational; customer bankruptcy; instability in markets; reduced capacity]}\\
{[environment, regulatory; climate change; regulatory developments; increase operating cost]}\\
\\
Test\\
\hline
\end{tabular}
\label{tab_semparseprompt}
\end{table}

\begin{table}[!htbp]
\caption{GPT4 prompt for generating text from semantic frames in an in-context learning setting.    }
\centering
\small
\begin{tabular}{|p{\linewidth}|}
\hline
Construct a text passage for the risk section of a SEC Filings report based on the given tuples. A tuple is in the format [<category>; <event>; <driver>; <impact>], where <category> is the risk category, <event> is the risk event (i.e. threat), <driver> is the underlying causes driving the risks, and <impact> is the impact/consequence. If a tuple item is n/a, then the corresponding value is not available.\\ 
\\
Example \#1\\
{[technology; changing technology; n/a; business suffers]}\\
{[operational; new digital products; complex and uncertain development process; no market acceptance]}\\

\\
Answer \#1\\
Our industry is characterized by rapidly changing technology and our success is dependent on our ability to adapt to such changes. Our business could suffer if we are unable to keep pace with rapid technological changes and product development in our industry. Our future success will largely depend on the successful introduction and sale of new digital products. Even if we successfully develop these products, the development of which is a complex and uncertain process requiring innovation and investment, they may not achieve market acceptance, which could have a material adverse effect on us.\\

\\
Test\\
\hline
\end{tabular}
\label{tab_genprompt}
\end{table}

\end{document}